%% file: main.tex
\documentclass[10pt,twocolumn,letterpaper]{article}

\usepackage{cvpr}              
\usepackage{multirow}
\usepackage{tablefootnote}  


\definecolor{cvprblue}{rgb}{0.21,0.49,0.74}
\usepackage[pagebackref,breaklinks,colorlinks,allcolors=cvprblue]{hyperref}

\def\paperID{*****} 
\def\confName{CVPR}
\def\confYear{2025}

\title{Logics-Parsing Technical Report}

\author{\textbf{Logics Team}\\
  Alibaba Group\\
{\tt\small (For the complete list of authors, please refer to the Contributors section.)}
}

\begin{document}
\maketitle
\input{sec/0_abstract}    
\input{sec/1_intro}

\input{sec/2_method}

\input{sec/3_experiments}

\input{sec/4_conclusion}
\input{sec/5_contribution}
{
    \small
    \bibliographystyle{ieeenat_fullname}
    \bibliography{main}
}


\end{document}

%% file: sec/0_abstract.tex
\begin{abstract}
Recent advances in Large Vision-Language models (LVLM) have spurred significant progress in document parsing task. Compared to traditional pipeline-based methods, end-to-end paradigms have shown their excellence in converting PDF images into structured outputs through integrated Optical Character Recognition (OCR), table recognition, mathematical formula recognition and so on. However, the absence of explicit analytical stages for document layouts and reading orders limits the LVLM's capability in handling complex document types such as multi-column newspapers or posters. To address this limitation, we propose in this report Logics-Parsing: an end-to-end LVLM-based model augmented with reinforcement learning. Our model incorporates meticulously designed reward mechanisms to optimize complex layout analysis and reading order inference. In addition, we expand the model's versatility by incorporating diverse data types such as chemical formulas and handwritten Chinese characters into supervised fine-tuning. Finally, to enable rigorous evaluation of our approach, we introduce LogicsParsingBench, a curated set of 1,078 page-level PDF images spanning nine major categories and over twenty sub-categories, which will be released later. Comprehensive experiments conducted on LogicsParsingBench have validated the efficacy and State-of-the-art (SOTA) performance of our proposed model across diverse document analysis scenarios.

\noindent\textbf{Project Page}: \href{https://github.com/alibaba/Logics-Parsing}{https://github.com/alibaba/Logics-Parsing}
\end{abstract}

%% file: sec/1_intro.tex
\section{Introduction}
\label{sec:intro}
Automatically parsing structured content from document PDF images has been a longstanding challenge in Document AI. The extracted elements including plain text, tables, formulas are foundational elements for advanced AI-based document analysis, Retrieval-augmented Generation (RAG) and ultimately the Artificial General Intelligence (AGI). Traditional OCR systems \cite{wang2024mineru,du2021pp} typically adopt pipeline-based architectures which sequentially deploy expert models to address sub-tasks including element detection, region cropping, element recognition and so on. However, this modular approach suffers from three key limitations: First, each expert model only optimizes its own local objectives, which does not guarantee a global optima for the entire document parsing task. Second, the computational overhead of training and deploying multiple models is substantial. Last but not least, the independent processing of multiple sub-components in one document disrupts inter-element information exchange and impairs holistic context modeling. To alleviate these drawbacks, end-to-end OCR-2.0 models \cite{wei2024general,wei2024vary} have been proposed to reduce maintenance costs and provide global optimization for various sub-tasks.

\begin{figure*}[htbp]
\centering
\includegraphics[width=\textwidth]{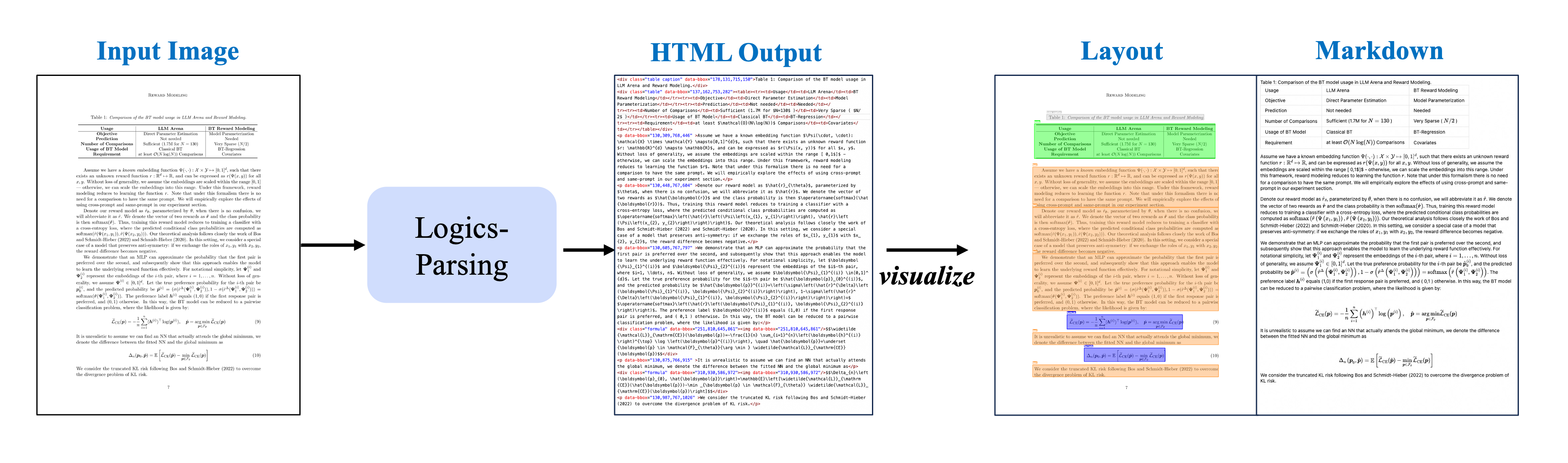}
\caption{The overview of our proposed Logics-Parsing for layout-aware document parsing.}
\label{fig:fig1}
\end{figure*}

In recent years, the rapid development of large vision-language models (LVLM) \cite{bai2023qwen,bai2025qwen2,liu2023visual,openai2023gpt} has largely improved the ability to handle diverse modalities of data. Despite the strong generalization ability, most LVLMs are primarily trained for reasoning tasks while exhibiting limitations for visual-text perception scenarios. Moreover, the CLIP objective \cite{radford2021learning} adopted by popular LVLMs' visual encoder like LLaVAR \cite{zhang2023llavar}, mPLUG-DocOwl \cite{ye2023mplug} and olmOCR \cite{poznanski2025olmocr} mainly focus on coarse-grained visual-text semantic alignment which neglects the fine-grained text information contained in images. Additionally, the pretrained clip vision models typically support fixed input resolutions (e.g., $224\times224$ or $336\times336$ pixels), which are
inadequate to meet the demands of parsing documents containing dense, small-sized text elements. To address these issues, Vary \cite{wei2024vary} has parallelly trained a high-resolution visual encoder specially for document-level OCR task. Qwen2-VL series \cite{wang2024qwen2,bai2025qwen2} have introduced the native dynamic resolution mechanism to eliminate the spatial constraints. Despite these improvements, end-to-end document parsing paradigms still face critical challenges. The next-token prediction objective widely adopted in current LVLMs naively optimizes the token-level alignment between input text images and output text, while overlooking the disruptive effects of complex layouts on reading order inference. To resolve this, Infinity-Parser \cite{wang2025infinity} introduced Reinforcement learning into the document parsing task with three layout-aware rewards. Despite the specially designed rewards, the paragraph count objective did not consider the explicit position and separation of corresponding element block by only calculating the number of paragraphs, and GRPO \cite{shao2024deepseekmath} performed directly on Qwen-2.5-VL-7B \cite{bai2025qwen2} base model does not ensure a prior foundation model for document parsing task, which may potentially mislead the direction of GRPO optimization. Recent study \cite{chu2025sftmemorizesrlgeneralizes} has highlighted a critical principle in post training of large-scale model: \textit{SFT memorizes and RL generalizes}, which emphasizes the importance of SFT for model output format stabilization before RL for generalization.

To address above limitations, in this report, we propose \textit{Logics-Parsing}, an end-to-end LVLM-based framework augmented with reinforcement learning to establish a robust and layout-aware document parsing model as shown in Fig.~\ref{fig:fig1}. A Two-stage SFT-then-RL training strategy is adopted to ensure the effectiveness of our model. Firstly, through incorporating diverse data types including normal texts, mathematical formulas, tables, chemical formulas and handwritten Chinese characters into training, we fine-tune the Qwen2.5-VL-7B \cite{bai2025qwen2} with more than 300K high-quality page-level document images to build a strong foundation for document parsing. Secondly, through layout-aware multi-component reward mechanisms together with a novel hard-sample mining strategy, we augment the model with the ability to correctly identify the page layout structure, then enforce adherence to natural reading sequences under the identified layout. This stage markedly improves performance on complex layouts such as multi-column newspapers and posters. In addition, in order to rigorously evaluate our proposed Logics-Parsing model, we construct a comprehensive benchmark containing 1,078 page-level PDF images spanning nine major categories (e.g., academic papers, technical reports) and over twenty sub-categories, called \textit{LogicsParsingBench}. Compared to OminiDocBench \cite{ouyang2025omnidocbench}, a top-leading benchmark covering diverse document types with rich annotation information, our proposed LogicsParsingBench places greater emphasis on assessing complex layout handling and scientific content parsing, with several enhanced evaluation protocols introduced.

In summary, the contributions of our work are listed as follows:
\begin{itemize}
\item  We propose Logics-Parsing, an end-to-end Large Vision-Language Model (LVLM)-based framework augmented with reinforcement learning. By explicitly modeling document layouts and supporting multimodal recognition of heterogeneous content types (e.g., text, mathematical formulas, tables), the framework enables the construction of a highly capable and broadly applicable model on complex, real-world document parsing scenarios.
\item To endow Logics-Parsing with robust recognition across diverse content types, we collected more than 300k high quality page-level document images and annotate them via an automated pipeline followed by human-in-the-loop verification. The co-occurrence of plain text, mathematical formulas, tables, and other elements on a single page offers comprehensive training for complex real-world document analysis.
\item Extensive evaluations on LogicsDocBench — a comprehensive benchmark comprising 1,078 page-level PDF images across nine categories — effectively validate that our method significantly improves structural understanding and content ordering in complex layouts, achieving superior performance in practical document parsing tasks. 
\end{itemize}

%% file: sec/2_method.tex
\section{Methodology}
\label{sec:method}

In this section, we present the methodology of Logics-Paring, with a two-stage training paradigm designed to establish outstanding capabilities for both document parsing and fine-grained structural alignment. We will first provide an overview of this SFT-then-RL framework, then describe the construction of high-quality training datasets, and finally detail each training stage.

\begin{figure}[htbp]
\centering
\includegraphics[width=0.45\textwidth]{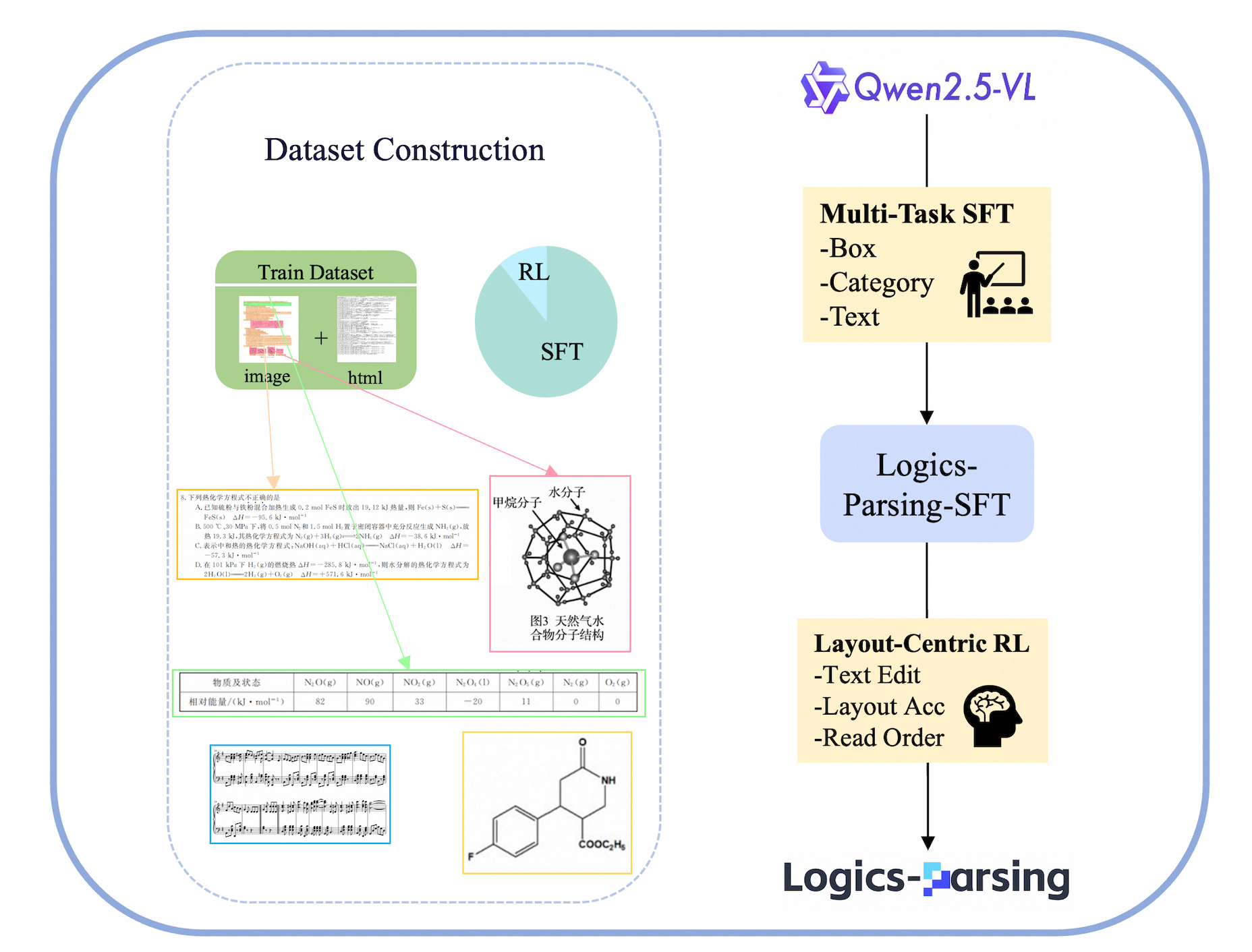}
\caption{The dataset construction and training pipeline of our proposed Logics-Parsing.}
\label{fig:fig2}
\end{figure}

\subsection{Overview}
As demonstrated in Fig.~\ref{fig:fig2}, the whole training process begins with a Supervised Fine-Tuning (SFT) stage. In this stage, the model is trained on a large-scale, versatile document dataset using a standard next-token prediction objective. This equips the model with essential capabilities such as text recognition and associating text blocks with spatial coordinates.

However, we argue that the autoregressive token-by-token supervision of SFT provides indirect and often insufficient signals for page structure learning. For example, the token-level cross-entropy loss does not explicitly penalize an incorrect paragraph-level reading order, nor does it differentiate between minor and major errors in bounding-box coordinates.

To address these limitations, the Layout-Centric Reinforcement Learning (LC-RL) is introduced. This stage leverages a smaller, curated set of challenging, high-quality samples for direct structual analysis optimization. By employing a multi-component reward function that explicitly evaluates text accuracy, layout precision, and logical reading order, we steer the model’s policy towards generating outputs that are not only textually accurate but also structurally coherent and logically sequenced.

\subsection{Dataset}
\label{sec:data}

To enable effective model training, we construct a large-scale, diverse, and high-quality dataset, with the HTML annotation format as one of its key features. As is known, HTML natively encodes the hierarchical structures of a document (e.g., nested tables, lists) and enables efficient rendering for visualization. The construction of this comprehensive dataset is designed to align with the two-stage SFT-then-RL training strategy, which consists of a SFT subset and a RL subset. This dual-subset design ensures the coherence between data preparation and corresponding model optimization phases.


\textbf{SFT Dataset.} To enrich training diversity and establish a strong foundational model for document parsing, we assembled data from two complementary sources.

On the one hand, we systematically incorporate and normalize multiple well-established public datasets into our unified HTML annotation schema. Specifically, for page-level parsing, we curate a subset of samples from olmOCR-mix-0225~\cite{poznanski2025olmocr}, systematically converting their original annotations into our standardized HTML format. For table recognition, which is a crucial sub-task, we collect data from leading benchmarks including FinTabNet~\cite{DBLP:journals/corr/abs-2005-00589}, TNCR~\cite{DBLP:journals/corr/abs-2106-15322}, and PubTabNet~\cite{DBLP:journals/corr/abs-1911-10683}. This has provided the model with extensive exposure to a wide variety of tabular structures.
For chemical structure recognition, we directly employ the ChEBI-20-MM dataset~\cite{liu2025quantitativeanalysisknowledgelearningpreferences}, a benchmark for molecular OCR. This offers a targeted training for interpreting the specific syntax and layout rules of chemical diagrams.

On the other hand, we build a large-scale in-house dataset annotated via a two-stage annotation pipeline consisting of automated extraction and expert model-based refinement. In the first place, Mathpix~\footnote{\url{https://mathpix.com/}} is employed to pre-annotate a broad corpus of public documents by extracting content, bounding boxes, and semantic types. Subsequently, to address complex layouts that Mathpix may struggle to parse, a verification and correction step is conducted using Gemini 2.5 Pro~\cite{comanici2025gemini}. This step focuses on correcting errors in challenging scenarios such as nested formulas and multi-layered tables, thereby ensuring the high accuracy and reliability of the annotations.

Nevertheless, there still exists a large number of highly complex documents and edge cases that the proposed automated annotation pipeline struggles to handle. For these samples, we introduce manual intervention, where approximately 10,000 page samples undergo a rigorous review and correction by expert human annotators. Apart from this, we further provide manual annotations on document reading order for these samples which remain challenging to achieve through automated tools. This human-in-the-loop approach has strictly guaranteed the annotation accuracy and broadens annotation coverage, yielding a gold-standard subset that significantly strengthens the model’s understanding of both content and structural organization in real-world layouts.

\textbf{RL Dataset.} Reinforcement learning usually benefits from a small but high-quality dataset. To this end, we propose a novel hard-sample mining strategy to identify a focused subset from the SFT corpus described above.

First, we manually curated around 4,000 pages from the above human-in-the loop subset. These samples are specifically chosen for their intricate and challenging layouts, which have been shown to present substantial challenges for standard autoregressive models.

Second, to collect samples that are \textit{difficult yet informative} for the RL stage learning, we employ an inference-guided hard-mining strategy. We run SFT model over the entire training set and select samples whose normalized edit distance between the prediction and the ground truth falls within a specific range $[0.5, 0.8]$. This range effectively captures instances that the SFT model may partially understand but fails to parse perfectly, forming an ideal "learning zone" for subsequent RL-based refinement. In this way, approximately 4,000 additional samples are collected.

Together, the above data-mining strategy has yielded roughly 8,000 high-quality, high-difficulty training samples for the LC-RL stage, whose effectiveness heavily hinges on a training set that specifically targets the weakness of the SFT model.

\subsection{Supervised Fine-Tuning}
\label{sec:sft}

The first stage of our training paradigm focuses on Supervised Fine-Tuning (SFT), designed to imbue the model with foundational document understanding capabilities and align the model output with the target structured HTML format.

Rather than training a large-scale Vision-Language Model (VLM) from scratch, we strategically choose to build our model upon a powerful and pre-existing foundation, where Qwen2.5-VL-7B-Instruct \cite{bai2025qwen2} is selected. This decision is driven by its state-of-the-art performance among open-source VLMs, which not only established a robust baseline for a wide range of vision-language tasks, but also enabled efficient adaptation for downstream domain-specific tasks. Specifically, the model is fine-tuned on the comprehensive dataset described in Section~\ref{sec:data} for a single epoch through the standard next-token prediction objective, from which it learns to generate the target HTML sequence conditioned on the input document image.

This process has yielded our baseline model called \textit{Logics-Parsing-SFT}. Upon completion of this stage, the model has already demonstrated remarkable proficiency in core document analysis tasks, which not only performs accurate text recognition and spatial grounding, but also exhibits a high degree of instruction-following capability, consistently generating structured HTML outputs aligned with prompt specifications. Despite a solid foundation model the Logics-Parsing-SFT has established, the inherent limitations of token-level supervision motivates a subsequent layout-centric reinforcement learning stage.

\subsection{Layout-centric Reinforcement Learning }
\label{sec:rl}

As illustrated above, the next-token prediction objective adopted in the SFT stage exhibits inherent limitations for complex layout and reading order learning, particularly in processing documents with highly intricate layouts such as newspapers, magazines, or multi-column reports. For these samples, parsing the correct reading order can be a non-trivial task even for humans. To overcome this challenge, we propose the second training phase based on Layout-centric Reinforcement Learning (LC-RL), employing Group Relative Policy Optimization (GRPO), which we find well-suited to this problem class. Through a comprehensive multi-component reward function that performs direct evaluation on the output answer quality and a carefully mined difficult-yet-informative dataset, we guide the model towards generating results that are more consistent with human reading habits.

Concretely, in each training iteration, we parse both the model's prediction and the ground-truth label to extract their constituent elements: textual content (including text, formulas, etc.) and their corresponding bounding box coordinates. These are then used to compute three distinct reward components. The first reward measures the character-level similarity between predicted text sequences and ground-truth ones through the negative normalized Levenshtein distance~\cite{lcvenshtcin1966binary}. The second reward is designed to evaluate the localization accuracy between the predicted and target bounding boxes, which guides the model to ground each content element at the correct page location. The last reward mainly pays attention on optimizing the logical reading flow of the parsed content, which is calculated as the pairwise inversion count between reference and predicted paragraphs. This reward directly penalizes out-of-order content and is crucial for learning complex, non-linear reading paths. Finally, we form the final reward as a linear combination of these three components for each sample.

\begin{figure*}[htbp]
\centering
\includegraphics[width=\textwidth]{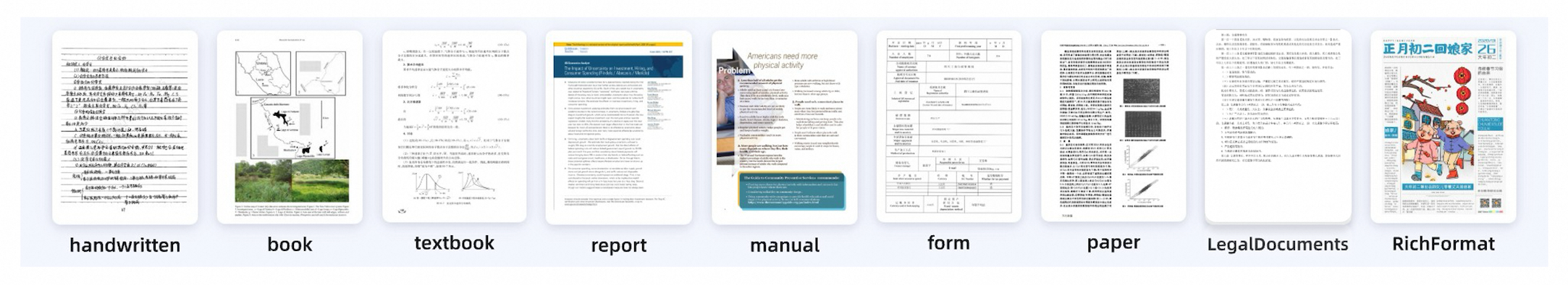}
\caption{The overview of document types included in our proposed LogicsParsingBench.}
\label{fig:fig3}
\end{figure*}

%% file: sec/3_experiments.tex
\section{Experiments}
\label{sec:exp}

\subsection{LogicsParsingBench: A Benchmark for Complex Document Parsing}
\label{sec:logicsdocbench}

To address the limitations of existing evaluation benchmarks, particularly their underrepresentation of documents with complex logical structures and specialized scientific content, we introduce LogicsParsingBench, a new challenging benchmark meticulously curated to rigorously assess the capabilities of modern document parsing models. Comprising 1,078 high-quality pages, LogicsParsingBench is designed to reflect real-world complexity and advance the frontiers of the document intelligence.

The benchmark is distinguished by three core features. Firstly, the samples in LogicsParsingBench span nine major categories specified in Fig.~\ref{fig:fig3} and over twenty sub-categories in both English and Chinese, which ensures diverse and comprehensive coverage. One critical category our proposed benchmark strongly emphasizes on is the academic and scientific documents sourced from a wide range of STEM fields (e.g., mathematics, physics, chemistry), characterized by high density of complex, nested mathematical formulas and intricate chemical structures. Additionally, rare and challenging document types such as music scroes and Chinese ancient books are also incorporated into the benchmark to further enrich the diversity of test samples.

Secondly, LogicsParsingBench includes a substantial portion of complex pages featuring multi-column layouts and text-graph mixed content, which not only imposes higher demands but also provides a more strict evaluation for model’s layout analysis and generalization capabilities. 

Last but not least, a more complex benchmark naturally necessitates a more nuanced evaluation protocol. Built upon the top-leading open-source evaluation code from OmniDocBench~\cite{ouyang2025omnidocbench}, the evaluation protocols in LogicsParsingBench feature two key improvements listed as follows:

\begin{itemize}
\item \textbf{Global Text Evaluation:} We claim that the commonly used block-by-block text matching evaluation would impose excessive penalty for minor differences in paragraph segmentation granularity. To mitigate this, we adopt a global text evaluation strategy, where all textual contents from one page (excluding headers and footers) are concatenated into a single string before calculating the Levenshtein distance. This provides a more holistic and robust measure for the core text recognition capability, invariant to the granularity of block segmentation.

\item \textbf{Stricter Content Normalization:} To ensure fair comparisons across heterogeneous model outputs, we implemented stricter normalization protocols, especially for tabular data. By eliminating redundant whitespace and simplifying specific LaTeX formatting tags, the evaluation would emphasize on semantic accuracy rather than superficial formatting differences.

\end{itemize}

\subsection{Implementation Details}

In our experiments, the Logics-Parsing model is built upon Qwen2.5-VL-7b-Instruct \cite{bai2025qwen2}, which is a strong foundation model for various multi-modality tasks. Leveraging the native dynamic resolution mechanism proposed in Qwen, we preserve the original aspect ratio of the input image while only constrain the total pixel count within $1024 \times 1024$ and the output token length within $8192$. This design balances the high-resolution input expectation of the task and the computation overhead. 

The model is trained in a two-stage manner. During the first SFT stage, the LLM component of Logics-Parsing is optimized to build foundational capabilities for document parsing, while the parameters of the vision encoder and vision-language projector are frozen. The model is optimized with a batch size of $256$ for $1$ epoch, and the learning rate is set to $2e-5$. During the second LC-RL stage, we transition to a reinforcement learning setup implemented via the ROLL~\cite{wang2025reinforcement}, adopting GRPO for policy optimization. For this fine-grained alignment, the hyperparameters are set more conservatively: the batch size is reduced to $32$, and the learning rate is lowered to $1e-6$. The model is trained for a fixed 250 steps on our curated RL dataset.

To enable a rigorous evaluation of our approach, we benchmark Logics-Parsing against state-of-the-art (SOTA) methods over the proposed LogicsParsingBench. Most evaluation metrics are aligned with OmniDocBench for consistency. Specifically, we calculate Normalized Edit Distance  (NED) \cite{levenshtein1965binary} between the predicted outputs and ground-truth references across all text components, including the pure texts, the HTML sequences for tabular data, the LaTeX expressions for mathematical formulas and the Simplified Molecular Input Line Entry System (SMILES) strings for chemical molecules. For reading order evaluation, we likewise report NED computed over text-only content, with tables, images, and ignored components excluded from the final reading order calculation.

\begin{figure}[htbp]
\centering
\includegraphics[width=0.45\textwidth]{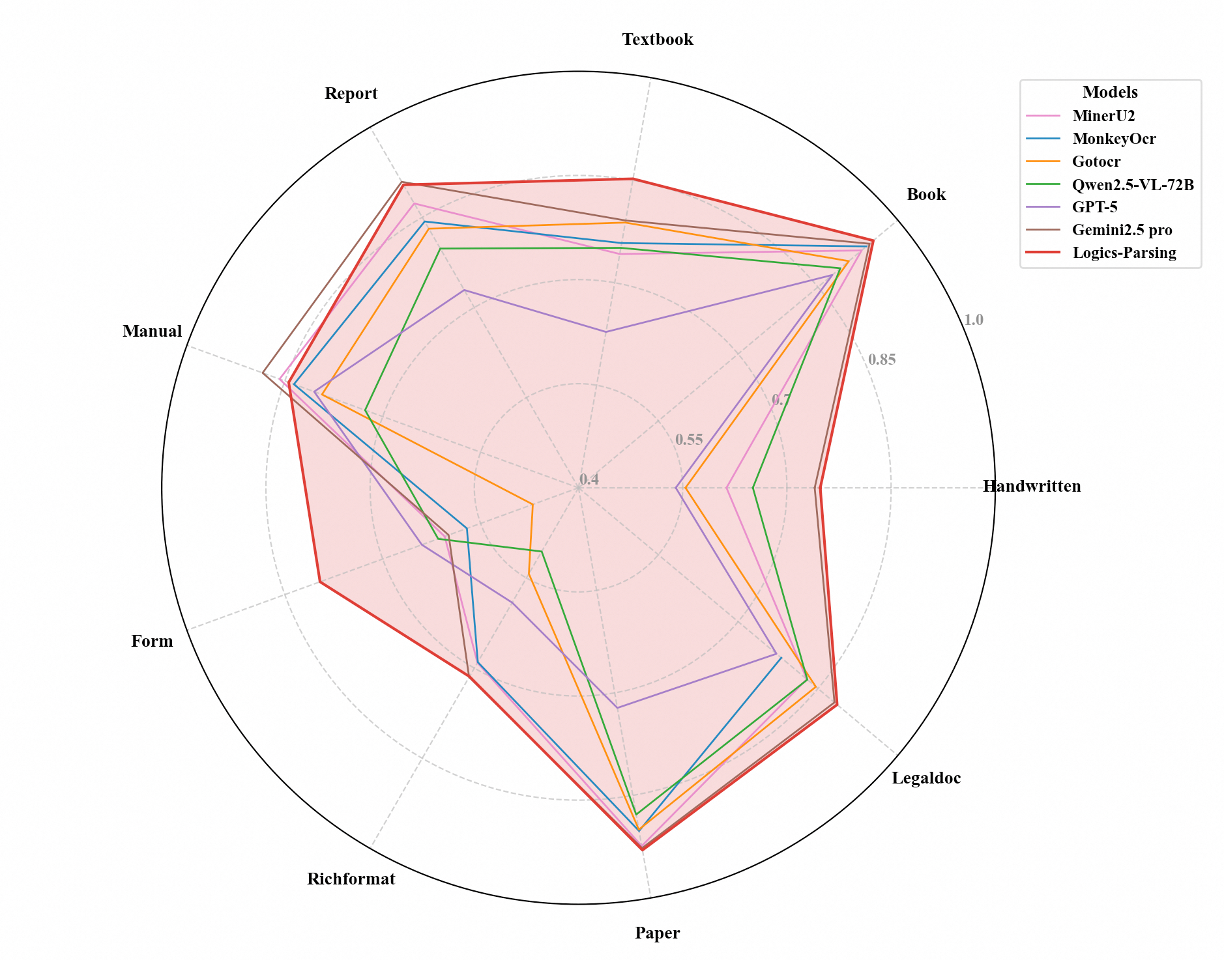}
\caption{Visualization of the comparisons between our propose Logics-Parsing and other methods across different document types.}
\label{fig:fig4}
\end{figure}

\begin{table*}[t]
\centering
\caption{Comparisons with State-of-the-art methods on LogicsParsingBench.}
\label{tab:logics-parsing-bench}
\resizebox{\textwidth}{!}{%
    \begin{tabular}{llcccccccccccccc}
    \toprule 
    \multirow{2}{*}{Model Type} & \multirow{2}{*}{Methods} & \multicolumn{2}{c}{Overall Edit $\downarrow$} & \multicolumn{2}{c}{Text Edit $\downarrow$} & \multicolumn{2}{c}{Formula Edit $\downarrow$} & \multicolumn{2}{c}{Table TEDS $\uparrow$} & \multicolumn{2}{c}{Table Edit $\downarrow$} & \multicolumn{2}{c}{ReadOrder Edit $\downarrow$} & \shortstack{Chemistry Edit $\downarrow$} & \shortstack{HW Edit $\downarrow$} \\
     & & EN & ZH & EN & ZH & EN & ZH & EN & ZH & EN & ZH & EN & ZH & ALL & ALL \\
    \midrule
    \multirow{7}{*}{Pipeline Tools} & doc2x~\footnotemark[2] & 0.209 & 0.188 & 0.128 & 0.194 & 0.377 & 0.321 & 81.1 & 85.3 & \underline{0.148} & \underline{0.115} & 0.146 & 0.122 & 1.0 & 0.307 \\
     & textin~\footnotemark[3] & 0.153 & 0.158 & 0.132 & 0.190 & 0.185 & 0.223 & 76.7 & \underline{86.3} & 0.176 & \textbf{0.113} & \textbf{0.118} & \textbf{0.104} & 1.0 & 0.344 \\
     & Mathpix~\footnotemark[1] & \underline{0.128} & \underline{0.146} & 0.128 & \underline{0.152} & \textbf{0.06} & \textbf{0.142} & \textbf{86.2} & \textbf{86.6} & \textbf{0.120} & 0.127 & 0.204 & 0.164 & 0.552 & \underline{0.253} \\
     & pp\_structure\_v3 \cite{cui2025paddleocr30technicalreport} & 0.220 & 0.226 & 0.172 & 0.29 & 0.272 & 0.276 & 66 & 71.5 & 0.237 & 0.193 & 0.201 & 0.143 & 1.0 & 0.382 \\
     & mineru2~\cite{wang2024mineru} & 0.212 & 0.245 & 0.134 & 0.195 & 0.280 & 0.407 & 67.5 & 71.8 & 0.228 & 0.203 & 0.205 & 0.177 & 1.0 & 0.387 \\
     & marker~\footnotemark[4] & 0.324 & 0.409 & 0.188 & 0.289 & 0.285 & 0.383 & 65.5 & 50.4 & 0.593 & 0.702 & 0.23 & 0.262 & 1.0 & 0.50 \\
     & pix2text~\footnotemark[5] & 0.447 & 0.547 & 0.485 & 0.577 & 0.312 & 0.465 & 64.7 & 63.0 & 0.566 & 0.613 & 0.424 & 0.534 & 1.0 & 0.95 \\
    \midrule
    \multirow{7}{*}{Expert VLMs} & Dolphin~\cite{feng2025dolphin} & 0.208 & 0.256 & 0.149 & 0.189 & 0.334 & 0.346 & 72.9 & 60.1 & 0.192 & 0.35 & 0.160 & 0.139 & 0.984 & 0.433 \\
     & dots.ocr~\footnotemark[6] & 0.186 & 0.198 & \underline{0.115} & 0.169 & 0.291 & 0.358 & 79.5 & 82.5 & 0.172 & 0.141 & 0.165 & 0.123 & 1.0 & 0.255 \\
     & MonkeyOcr~\cite{li2025monkeyocrdocumentparsingstructurerecognitionrelation} & 0.193 & 0.259 & 0.127 & 0.236 & 0.262 & 0.325 & 78.4 & 74.7 & 0.186 & 0.294 & 0.197 & 0.180 & 1.0 & 0.623 \\
     & OCRFlux~\footnotemark[7] & 0.252 & 0.254 & 0.134 & 0.195 & 0.326 & 0.405 & 58.3 & 70.2 & 0.358 & 0.260 & 0.191 & 0.156 & 1.0 & 0.284 \\
     & gotocr~\cite{wei2024general} & 0.247 & 0.249 & 0.181 & 0.213 & 0.231 & 0.318 & 59.5 & 74.7 & 0.38 & 0.299 & 0.195 & 0.164 & 0.969 & 0.446 \\
     & olmocr~\cite{poznanski2025olmocr} & 0.341 & 0.382 & 0.125 & 0.205 & 0.719 & 0.766 & 57.1 & 56.6 & 0.327 & 0.389 & 0.191 & 0.169 & 1.0 & 0.294 \\
     & SmolDocling~\cite{nassar2025smoldoclingultracompactvisionlanguagemodel} & 0.657 & 0.895 & 0.486 & 0.932 & 0.859 & 0.972 & 18.5 & 1.5 & 0.86 & 0.98 & 0.413 & 0.695 & 1.0 & 0.927 \\
    \midrule
    \multirow{5}{*}{General VLMs} & Qwen2-VL-72B~\cite{wang2024qwen2} & 0.298 & 0.342 & 0.142 & 0.244 & 0.431 & 0.363 & 64.2 & 55.5 & 0.425 & 0.581 & 0.193 & 0.182 & 0.792 & 0.359 \\
     & Qwen2.5-VL-72B~\cite{bai2025qwen2} & 0.233 & 0.263 & 0.162 & 0.24 & 0.251 & 0.257 & 69.6 & 67 & 0.313 & 0.353 & 0.205 & 0.204 & 0.597 & 0.349 \\
     & doubao-1.6~\footnotemark[8] & 0.188 & 0.248 & 0.129 & 0.219 & 0.273 & 0.336 & 74.9 & 69.7 & 0.180 & 0.288 & 0.171 & 0.148 & 0.601 & 0.317 \\
     & GPT-5~\footnotemark[9] & 0.242 & 0.373 & 0.119 & 0.36 & 0.398 & 0.456 & 67.9 & 55.8 & 0.26 & 0.397 & 0.191 & 0.28 & 0.88 & 0.46 \\
     & Gemini2.5 pro~\cite{comanici2025gemini} & 0.185 & 0.20 & \underline{0.115} & 0.155 & 0.288 & 0.326 & \underline{82.6} & 80.3 & 0.154 & 0.182 & 0.181 & 0.136 & 0.535 & 0.26 \\
     \midrule
     Expert VLMs & \shortstack{Logics-Parsing} & \textbf{0.124} & \textbf{0.145} & \textbf{0.089} & \textbf{0.139} & \underline{0.106} & \underline{0.165} & 76.6 & 79.5 & 0.165 & 0.166 & \underline{0.136} & \underline{0.113} & \textbf{0.519} & \textbf{0.252} \\
    \bottomrule
    \end{tabular}%
}

\begin{minipage}{\textwidth}
\vspace{1ex} 
\footnotesize
\textit{Note:} Bold text indicates the best result, and underlined text indicates the second-best result. 
\end{minipage}
\end{table*}

\subsection{Main Results}
\subsubsection{Comparison with State-of-the-Art Methods}
In this section, we report the experimental results of our proposed Logics-Parsing against a wide range of existing document parsing models in Tab.~\ref{tab:logics-parsing-bench}, including state-of-the-art open-source models~\cite{cui2025paddleocr30technicalreport,feng2025dolphin,wei2024general,bai2025qwen2} as well as leading commercial tools like doc2x~\footnote{\url{https://doc2x.noedgeai.com/}} and TextIn~\footnote{\url{https://www.textin.com}}. Fig.~\ref{fig:fig4} also provides a visualized comparison on the performance across multiple document types, which gives a more intuitive yet comprehensive demonstration on the superiority and well-balanced capability of our proposed Logics-Parsing against other state-of-the-art methods.

Specifically, as is shown in Tab.~\ref{tab:logics-parsing-bench}, we categorize existing methods into three groups, namely Pipeline Tools, Expert VLMs and General VLMs, which collectively represent the mainstream technical paradigms in the field of document parsing. Our Logics-Parsing falls within the Expert VLMs category, which aims to build a professional task-specific model for document parsing. The quantitative results have demonstrated that Logics-Parsing achieves the State-of-the-Art (SOTA) performance overall, attaining the lowest aggregate edit distance on both English ($0.124$) and Chinese ($0.145$) documents. A detailed breakdown has further revealed Logics-Parsing's superior capabilities over all other methods in parsing pure text, chemical structures, and handwritten content. 

~\footnotetext[4]{\url{https://github.com/datalab-to/marker}}
~\footnotetext[5]{\url{https://github.com/breezedeus/Pix2Text}}
~\footnotetext[6]{\url{https://github.com/rednote-hilab/dots.ocr}}
~\footnotetext[7]{\url{https://ocrflux.pdfparser.io/}}
~\footnotetext[8]{\url{https://seed.bytedance.com/zh/seed1_6}}
~\footnotetext[9]{\url{https://openai.com/zh-Hans-CN/index/introducing-gpt-5/}}

\begin{table*}[htbp]
\centering
\caption{Ablation study on the effectiveness of the SFT-then-RL paradigm.}
\label{tab:ablation_grpo}
\resizebox{\textwidth}{!}{%
\begin{tabular}{lcccccccccccccc}
\toprule
\multirow{2}{*}{Methods} & \multicolumn{2}{c}{Overall Edit ↓} & \multicolumn{2}{c}{Text Edit ↓} & \multicolumn{2}{c}{Formula Edit ↓} & \multicolumn{2}{c}{Table TEDS ↑} & \multicolumn{2}{c}{Table Edit ↓} & \multicolumn{2}{c}{ReadOrder Edit ↓} & Chemistry Edit ↓ & HW Edit ↓\\
 & EN & ZH & EN & ZH & EN & ZH & EN & ZH & EN & ZH & EN & ZH & ALL & ALL \\
\midrule
Qwen2.5-VL-7B(baseline) & 0.316 & 0.319 & 0.149 & 0.185 & 0.33 & 0.302 & 64.5 & 63.3 & 0.585 & 0.637 & 0.202 & 0.152 & 0.75 & 0.337 \\
Logics-Parsing-SFT & 0.133 & 0.159 & 0.093 & 0.165 & 0.113 & 0.169 & 74.1 & \textbf{80.32} & 0.176 & \textbf{0.149} & 0.149 & 0.155 & 0.520 & 0.276 \\
Logics-Parsing & \textbf{0.124} & \textbf{0.145} & \textbf{0.089} & \textbf{0.139} & \textbf{0.106} & \textbf{0.165} & \textbf{76.6} & 79.5 & \textbf{0.165} & 0.166 & \textbf{0.136} & \textbf{0.113} & \textbf{0.519} & \textbf{0.252} \\
\bottomrule
\end{tabular}
}
\end{table*}

\begin{figure*}[htbp]
\centering
\includegraphics[width=\textwidth]{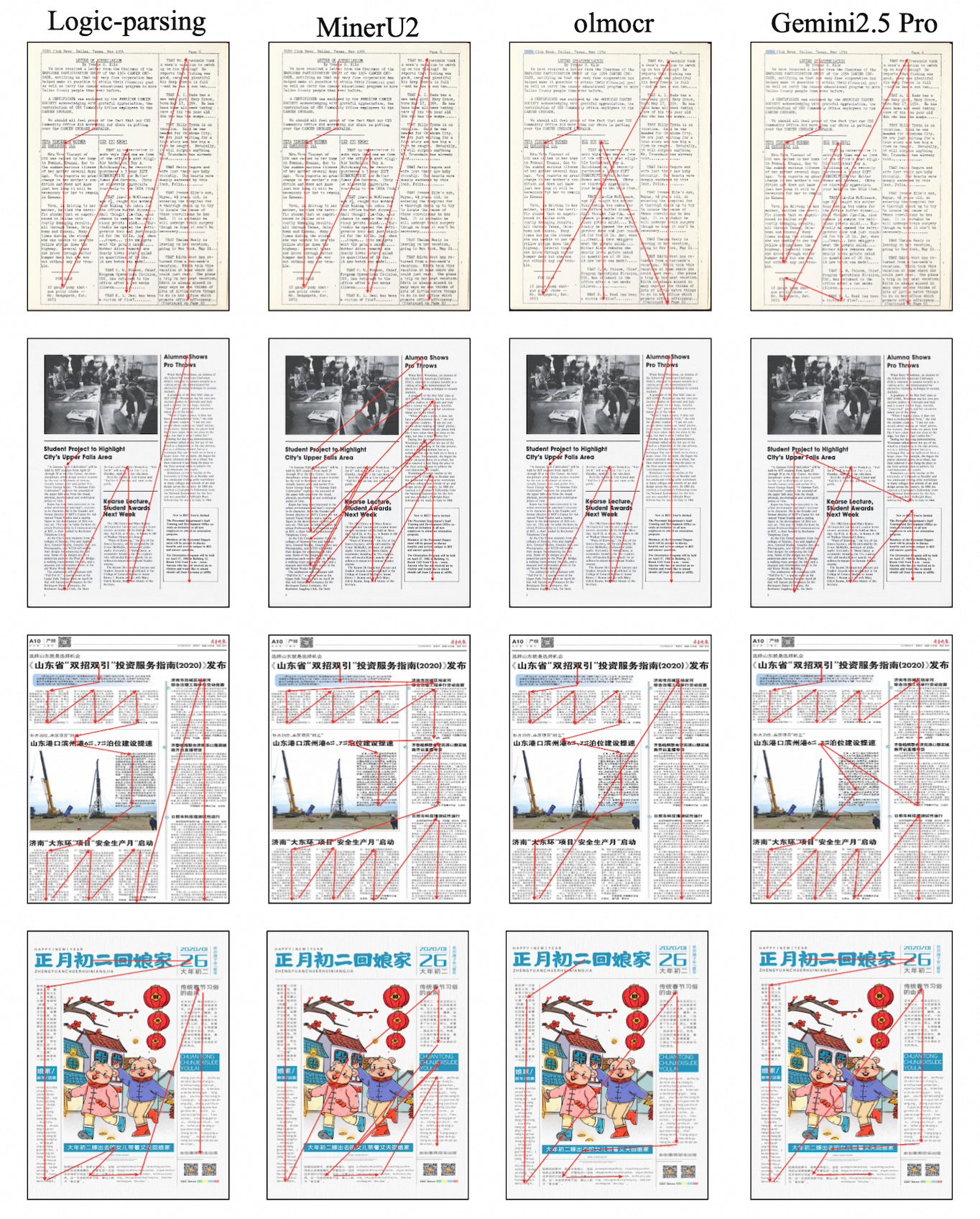}
\caption{Qualitative comparison of reading order prediction. Red arrows denote the predicted flow. Our Logics-Parsing model generates a reading sequence fully aligned with the ground-truth.}
\label{fig:fig5}
\end{figure*}

Despite an inferior performance in reading order prediction compared with the commercial tool TextIn, our proposed Logics-Parsing significantly surpasses all other open-source alternatives. Some qualitative comparisons of reading order prediction are presented in Fig.~\ref{fig:fig5}. The results show that the reading orders predicted by our Logics-Parsing are both visually clear and structurally coherent, closely following the standard left-to-right and top-to-bottom reading conventions, while accurately preserving contextual relationships among text paragraphs and semantic regions.

Moreover, while Logics-Parsing has demonstrated superior performance across most dimensions, we acknowledge a performance gap existed in table structure recognition and mathmetic formula recognition compared to the top-leading methods. These shortcomings highlight a direction for future improvement, particularly in modeling complex two-dimensional layouts.

\subsubsection{Ablation Study}
We perform ablation experiments to validate the effectiveness of our proposed SFT-then-RL framework. We explicitly compare two models, one of which is the Logics-Parsing-SFT defined in Sec.~\ref{sec:sft}, another one is the final Logics-Parsing model which is augmented with layout-centric reinforcement learning.

The experimental results are presented in Tab.~\ref{tab:ablation_grpo}. We observe that after fine-tuning on self-constructed SFT dataset, the Logics-Parsing-SFT model has developed remarkable document parsing abilities across different content types compared with the original Qwen2.5-VL-7B baseline. This performance provides strong validation on the high quality of both the collected training data and the annotations generated through a combination of automated and human-in-the-loop manner. After the second LC-RL stage, the final Logics-Parsing model consistently outperforms the Logics-Parsing-SFT model across nearly all metrics for both English (EN) and Chinese (ZH). The most significant improvement appears on the ReadOrder Edit distance, which our proposed multi-component reward mechanism mainly optimizes on. Specifically, the ReadOrder score for English drops from $0.149$ to $0.136$, and for Chinese, a substantial reduction from $0.155$ to $0.113$ is observed. This phenomenon has provided strong evidence that fine-tuning on a small, curated set of explicitly mined hard samples with our proposed layout-centric GRPO strategy effectively enhances the model's ability to handle complex document layouts and inference accurate reading order.

%% file: sec/4_conclusion.tex
\section{Conclusion}
\label{sec:conclusion}

We present Logics-Parsing, an end-to-end LVLM-based document parsing model augmented with layout-centric reinforcement learning. Through a specially designed two-stage SFT-then-RL training framework and carefully curated training datasets tailored to distinct characteristics of each stage, Logics-Parsing has equipped strong abilities for both explicit document layout modeling and multimodal recognition of heterogeneous content types.

To rigorously evaluate our proposed framework, we proposed a new benchmark called LogicsParsingBench consisting of 1,078 page-level high-quality PDF images across diverse categories. With more diverse and complex document types such as academic papers included and improved evaluation protocols compared to OminiDocBench, LogicsParsingBench offers a more fair and challenging benchmark reflecting real-world document complexity and advancing the frontiers of the document intelligence research.

In the future, we will pay more attention to explore architectural innovation for the SFT training and develop more focused and fine-grained reward mechanism for improved layout-centric reinforcement learning. These directions hold promising advancement toward a document parsing model that is highly versatile yet deeply specialized, capable of professionally handling a wide array of document understanding tasks with greater accuracy and adaptability.

%% file: sec/5_contribution.tex
\section{Contributors}
\label{sec:contribution}

\noindent Xiangyang Chen, Shuzhao Li, Xiuwen Zhu, Yongfan Chen, Fan Yang, Cheng Fang, Lin Qu, Xiaoxiao Xu, Hu Wei, Minggang Wu.